\pdfoutput=1

\documentclass[11pt]{article}

\usepackage{EMNLP2022}

\usepackage{times}
\usepackage{latexsym}

\usepackage[T1]{fontenc}

\usepackage[utf8]{inputenc}

\usepackage{microtype}

\usepackage{inconsolata}

\usepackage{algorithmicx}
\usepackage{algorithm}
\usepackage{graphicx}
\usepackage{amsmath}
\usepackage{amsthm}
\usepackage{booktabs}
\usepackage{multicol}  
\usepackage{multirow}
\usepackage{mathrsfs}
\usepackage{color}
\usepackage{array}

\usepackage{booktabs}
\usepackage{amssymb}
\usepackage{lscape}
\usepackage{ragged2e}
\usepackage{enumerate}
\usepackage{enumitem}
\usepackage{subfigure}

\usepackage[nameinlink]{cleveref}

\crefformat{section}{\S#2#1#3} 
\crefname{algorithm}{Alg.}{Algs.}
\crefformat{subsection}{\S#2#1#3}
\Crefname{equation}{Eq.}{Eqs.}
\Crefname{figure}{Fig.}{Figs.}

\title{Alleviating Sparsity of Open Knowledge Graphs \\ with Ternary Contrastive Learning}

\author{Qian Li\textsuperscript{\rm 1,2}, Shafiq Joty\textsuperscript{\rm 2,3}, Daling Wang\textsuperscript{\rm 1}\thanks{$^*$Corresponding author} , Shi Feng\textsuperscript{\rm 1} and Yifei Zhang\textsuperscript{\rm 1} \\
\textsuperscript{\rm 1} Northeastern University, China\\
\textsuperscript{\rm 2} Nanyang Technological University, Singapore\\
$^3$Salesforce Research\\
\texttt{feiwangyuzhou@foxmail.com, srjoty@ntu.edu.sg}\\
\texttt{\{wangdaling,fengshi,zhangyifei\}@cse.neu.edu.cn}
}

\begin{document}
\maketitle
\begin{abstract}
Sparsity of formal knowledge and roughness of non-ontological construction make sparsity problem particularly prominent in Open Knowledge Graphs (OpenKGs). Due to sparse links, learning effective representation for few-shot entities becomes difficult. We hypothesize that by introducing negative samples, a contrastive learning (CL) formulation could be beneficial in such scenarios. However, existing CL methods model KG triplets as binary objects of entities ignoring the relation-guided ternary propagation patterns and they are too generic, i.e., they ignore zero-shot, few-shot and synonymity problems that appear in OpenKGs. To address this, we propose TernaryCL, a CL framework based on ternary propagation patterns among head, relation and tail. TernaryCL designs  \emph{Contrastive Entity} and \emph{Contrastive Relation} to mine ternary discriminative features with both negative entities and relations, introduces \emph{Contrastive Self} to help zero- and few-shot entities learn discriminative features, \emph{Contrastive Synonym} to model synonymous entities, and \emph{Contrastive Fusion} to aggregate graph features from multiple paths. Extensive experiments on benchmarks demonstrate the superiority of TernaryCL over state-of-the-art models. 
  

\end{abstract}

\section{Introduction}

\begin{figure*}
	\centering
	\includegraphics[width=2.0\columnwidth]{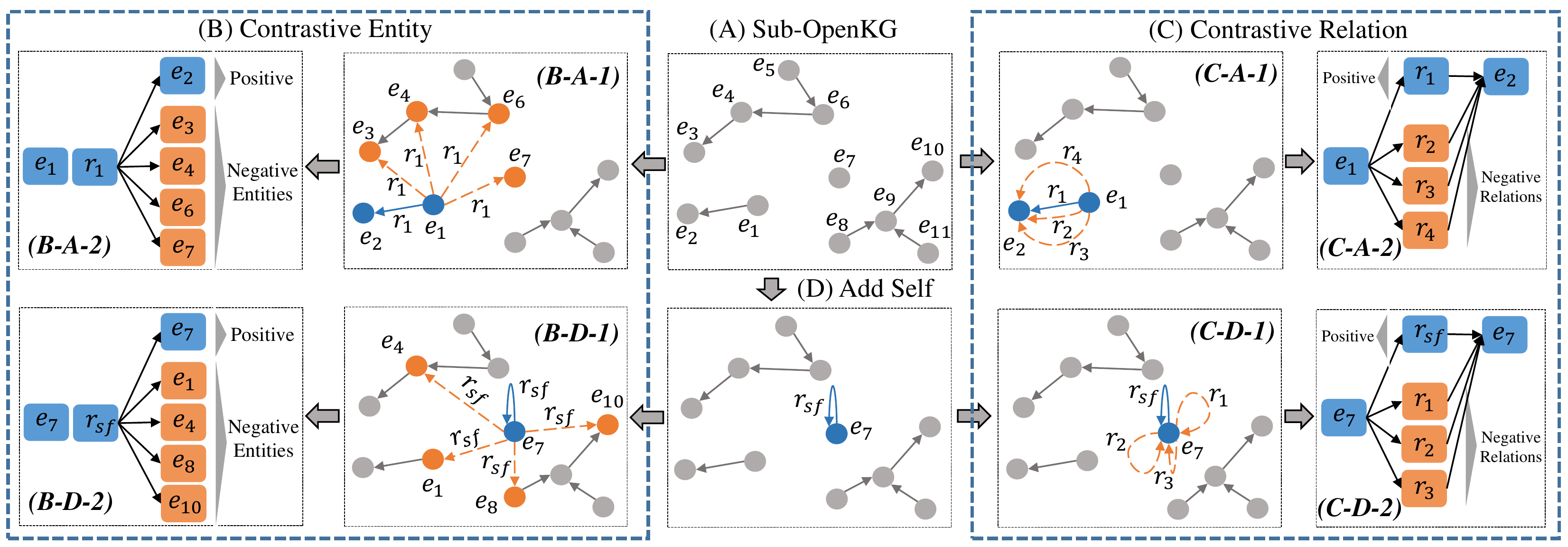}
	\caption{Framework of TernaryCL for alleviating sparsity of OpenKGs. (A) A given sub-OpenKG. (B) Contrastive Entity generates negative entities (yellow) and contrasts them with a positive entity (blue).
	(C) Contrastive Relation generates negative relations (yellow) and contrasts them with a positive relation (blue).
	(D) Contrastive Self constructs a positive sample by adding a \emph{self} relation $r_{sf}$ (blue) to the entity (blue), generates negative entities (yellow) in (B-D) and negative relations (yellow) in (C-D), and contrasts them with the \emph{self} positive sample.
	}
	\label{model}
\end{figure*}


Open Knowledge Graphs (OpenKGs) structure textual facts in the form of (\emph{subject entity}, \emph{relation}, \emph{object entity}) without depending on an ontology schema \cite{FaderSE11,GashteovskiWHBG19}. They benefit knowledge-intensive tasks, such as question answering \cite{SunBC19} and dialogue systems \cite{DinanRSFAW19}. 
Representation learning of OpenKGs aims to learn implicit embeddings of entities and relations \cite{GuptaKT19,BroscheitGWG20}, and has become an indispensable step in the applications of OpenKGs. 



Due to the rigidness of grammatical patterns and roughness of non-ontological construction, a common challenge in representation learning of OpenKGs is the sparsity problem, where a large portion of entities have few- or zero-shot links. In the two standard OpenKGs: ReVerb20K and ReVerb45K \cite{VashishthJT18}, the degree of $55\%$ and $89\%$ entities is less than $3$.
As such, few- and zero-shot entities in these OpenKGs do not get enough training, resulting in poor generalization.




We hypothesize that negative samples could help existing sparse links to learn discriminative features in the form of a negative feedback.
Being popular in self-supervised learning, \emph{contrastive learning} (CL) aims to learn representations by contrasting negative samples with the positive ones \cite{He0WXG20,SimCSE,0001XYLWW21}. Although existing CL in Graphs \cite{VelickovicFHLBH19} and CuratedKGs \cite{AhrabianFSHB20,0046ZWL22} have shown promising results, they could not effectively tackle the sparsity and synonymity problems of OpenKGs. First, they model KG triplets as binary objects of entities, ignoring the relation-guided ternary propagation patterns where entities propagate to multiple neighbor entities through multiple relations. Second, they do not specifically tackle the sparsity issues of zero- and few-shot entities. Finally, they can not address the synonymity problem in OpenKGs where multiple entities with different surface forms have the same meaning.





To alleviate the above problems, we propose TernaryCL, a contrastive learning framework based on ternary relational patterns among head, relation and tail. TernaryCL uses the following key ideas: 
(1) {\textbf{Contrastive Entity}} learns ternary discriminative features of different entities under the same (head entity, relation)-pair, which alleviates the sparsity issue by negative entities (\Cref{model}B); (2) \textbf{Contrastive Relation} learns ternary discriminative features of different relations under the same (head entity, tail entity)-pair, which alleviates the sparsity by negative relations (\Cref{model}C);
(3) \textbf{Contrastive Self} designs \emph{self} positive samples to give zero- and few-shot entities chances to learn discriminative features (\Cref{model}D);
(4) \textbf{Contrastive Synonym} constructs \emph{synonymous} positive samples to automatically aggregate synonymous entities (\Cref{model_syn}); 
(5) \textbf{Contrastive Fusion}  extends the ternary propagation pattern from the above 1-to-1 to 1-to-N: a head entity propagates to multiple tail entities through multiple relations. To gain insights into the method, we analyze the gradients of the above components.


We show that TernaryCL can learn effective embeddings from the KG itself without relying on pretrained language models or external information. This makes it light-scale and easy to apply to downstream tasks. We perform extensive evaluation to understand its performance on full-shot and at different sparsity levels, and for few- and zero-shot types. Our results show that TernaryCL can significantly outperform state-of-the-art baselines. 

In summary, our key contributions are:
\begin{itemize}[leftmargin=*]
	\item To the best of our knowledge, this is the first work to do contrastive learning
	with relation and synonymous views over OpenKGs.
	\item We propose a ternary contrastive learning model, TernaryCL, to learn representations from complex propagation patterns of OpenKGs.
	It subtly generates negative samples from the perspective of entities and relations, and uses contrastive learning to incorporate structural information. It does not rely on external resources making it easy to scale and to apply to downstream tasks.   
	\item We introduce \emph{Contrastive Self} to solve the problem of learning for zero- and few-shot entities, and  propose \emph{Contrastive Synonym} to aggregate signals from synonymous entities. 
	\item We perform extensive experiments to show the superiority of our method over state-of-the-art baselines. We release our code at \url{https://github.com/feiwangyuzhou/TernaryCL}.
\end{itemize}


\section{Related Work}
OpenKGs represent factual knowledge in structured forms, as triples of \emph{h}ead-\emph{r}elation-\emph{t}ail or $(h, r, t)$. They are extracted with OpenIE tools \cite{FaderSE11,GashteovskiWHBG19}, and generally do not rely on specification of ontology.
Although OpenKGs have the advantage that they can be easily bootstrapped to new domains, because of the sparsity of formal grammatical patterns and  non-ontological construction, many relevant facts are often missing from such OpenKGs. This makes them difficult to be used effectively in downstream tasks \cite{ChandrahasT21}. 

Representation learning of KGs devotes to learning informative features of entities and relations, which can be used for other KG-related downstream tasks.
General representation learning models over KGs focus on inducing structural features with linear \cite{bordes2013translating}, bilinear \cite{wang2014knowledge,lin2015learning}, complex \cite{yang2014embedding,trouillon2016complex} or convolutional \cite{dettmers2018convolutional,nguyen2017novel} operations, while OpenKG-specific models enhance the embeddings with side information \cite{GuptaKT19} and pretrained language models \cite{ChandrahasT21}.
However, these methods are limited in alleviating the sparsity issue. We propose a contrastive learning method that is more effective and does not rely on external resources.

Contrastive learning aims to learn effective representation by pulling close neighbors and pushing apart non-neighbors \cite{HadsellCL06,SimCSE}, 
which has achieved great success in vision \cite{He0WXG20}, text \cite{SimCSE} and graph \cite{0001XYLWW21}. Contrastive learning methods in Graphs attempt to leverage a contrastive loss at node \cite{VelickovicFHLBH19}, graph \cite{SunHV020} and multi-view levels \cite{HassaniA20,0001XYLWW21}.
\citet{AhrabianFSHB20} and \citet{0046ZWL22} introduce contrastive learning into CuratedKGs by designing negative sampling strategies.
However, existing contrastive learning on Graphs and CuratedKGs only model binary objects. In contrast, we propose contrastive learning to study ternary patterns in OpenKGs.

%



\section{Preliminaries}
\label{preliminaries}


$\bullet$ Let $\mathcal{G} = (\mathcal{E},$ $\mathcal{R})$ be an \textbf{OpenKG} and ($h,r,t$) be a triple in $\mathcal{G}$ with $h, t$ $\in$ $\mathcal{E}$ being the head and tail entities, and $r$ $\in$ $\mathcal{R}$ being the relation between them.
Entities and relations are non-empty word sequences; let $w_h = \{ w_{h,i} \}_{i=1}^{|w_h|}$ and $w_r = \{ w_{r,i} \}_{i=1}^{|w_r|}$ represent word sequences of entity $h$ and relation $r$ respectively.
Representations of entities and relations are denoted as $\mathbf{E}$ $\in$ $\mathbb{R}^{|\mathcal{E}| \times D}$ and $\mathbf{R}$ $\in$ $\mathbb{R}^{|\mathcal{R}| \times D}$ respectively, with $D$ being the embedding dimension.
We will use $\mathbf{h}$ $\in$ $\mathbf{E}$ and $\mathbf{r}$ $\in$ $\mathbf{R}$ to denote the embeddings of entity $h$ and relation $r$, respectively.

\noindent $\bullet$ \textbf{Link prediction} is used as a downstream task to verify the effectiveness of representation learning.
Link prediction in OpenKGs is to predict the answers for the two questions: (1) predicting the tail $Q_{t}$=$(h, r, ?)$ and (2) predicting the head $Q_{h}$=$(?, r, t)$. 
For each such question, the possible answer entities can be one or more, because there could be multiple entities with the same meaning but different textual forms in an OpenKG \cite{BroscheitGWG20}.
For example, for question $Q_{t}$=(\emph{“NBC-TV”}, \emph{“has office in”}, ?), we expect all answers from the set of entities \{\emph{“New York”, “NYC”, “New York City”}\}.


\noindent $\bullet$ A \textbf{zero-shot} entity (relation) is an entity (relation) without any link in the KG.
A \textbf{few-shot} entity (relation) is an entity (relation) with few links in the KG, e.g., an one-shot entity has only one link.
A zero-shot (few-shot) triple is a triple that contains at least one zero-shot (few-shot) entity or relation.



\section{Proposed TernaryCL Model}


We propose TernaryCL to alleviate the sparsity of OpenKGs in representation learning (\Cref{model}).
In the following, we first introduce a simple \textit{Ternary Similarity} function to compute a similarity score for each $(h,r,t)$ triple in an OpenKG by considering both textual and structural information (\cref{subsec:emb}). We present \textit{Contrastive Entity} in \Cref{subsec:CE} to learn embeddings of different entities with the same $(h, r)$-pair, followed by \textit{Contrastive Relation} in \Cref{subsec:CR} to learn embeddings of different relations with the same $(h, t)$-pair. We describe \textit{Contrastive Self} in \Cref{subsec:CSF} that constructs a positive sample ($h, r_{sf}, h^{+}$) and contrasts it with negative samples to give zero- and few-shot entities chances to learn informative features. We present \textit{Contrastive Synonym} in \Cref{subsec:CSM} to aggregate signals from multiple synonymous entities.  \textit{Contrastive Fusion} in \Cref{subsec:CF} futher extends propagation patterns from the above 1-1 to 1-N.
Finally, the training procedure is described in \Cref{subsec:train}.

\subsection{Ternary Similarity} \label{subsec:emb}




For a triple $(h,r,t) \in \mathcal{G}$, word sequence of head entity $h$ is 
$w_h = \{ w_{h,i} \}_{i=1}^{|w_h|}$,
of relation $r$ is 
$w_r = \{ w_{r,i} \}_{i=1}^{|w_r|}$, and of tail entity $t$ is 
$w_t = \{ w_{t,i}  \}_{i=1}^{|w_t|}$. 
We encode each of these sequences with a text encoder ($\text{Enc}$) such as  
BiGRU \cite{cho2014learning} and BERT \cite{DevlinCLT19}.
\begin{equation}
	\mathbf{i^w} = \text{Enc}(w_i) \text{ for } w_i \in \{w_h, w_r, w_t \}
	\label{bigru}
\end{equation}
This  yields the \emph{textual} embeddings of $h$, $r$ and $t$ as $\mathbf{h^w}$, $\mathbf{r^w}$ and $\mathbf{t^w}$, respectively. For BiGRU, we concatenate the last states in the forward and backward directions to get the sequence representation. For BERT, we take the [CLS] representation.


Then, we focus on exploiting potential connections between entities and relations.
We use a two-dimensional convolutional network \citep{dettmers2018convolutional} to learn potential connections between a head entity $h$ and a relation $r$ as follows: 
\begin{equation}
	\varphi (h,r) = \rho\big( \text{Linear}(\rho(\text{Conv2d}_{\omega}([\widehat{\mathbf{h}};\widehat{\mathbf{r}}]))) \big) \label{eq:enc}
\end{equation} 
where $\rho$ represents a ReLU activation, and $\widehat{\mathbf{h}}$ and $\widehat{\mathbf{r}}$ denote a reshaping of $[\mathbf{h} + \mathbf{h^w}]$ and $\mathbf{r^w}$ respectively, with $\mathbf{h} \in \mathbf{E}$.\footnote{Recall that $\mathbf{h} \in \mathbf{E}$ and $\mathbf{r} \in \mathbf{R}$ are the embedding parameters of entity $h$ and relation $r$, which are initialized randomly and updated during training. For relations, adding $\mathbf{r} \in \mathbf{R}$ with the textual embedding $\mathbf{r^w}$ worsen the performance.} 
The reshaping operation converts a vector $\mathbf{v} \in \mathbb{R}^D$ from one-dimension $D$ to two-dimensions $\mathbf{v} \in \mathbb{R}^{D_1 \times D_2}$, where $D=D_1.D_2$; $[\widehat{\mathbf{h}};\widehat{\mathbf{r}}] \in \mathbb{R}^{(2D_1) \times D_2}$ represents the concatenation of the reshaped embeddings of $\widehat{\mathbf{h}}$ and $\widehat{\mathbf{r}}$. The $\text{Conv2d}_{\omega}$ symbol denotes a two-dimensional convolutional layer with filters $\omega$. This layer returns a feature map tensor $\mathcal{C} \in \mathbb{R}^{C_1 \times C_2 \times C_3}$, where $C_1$ is the number of feature maps of dimensions $C_2 \times C_3$. $\mathcal{C}$ is then reshaped into a vector $\mathbb{R}^{C_1.C_2.C_3}$, and projected back to $\mathbb{R}^D$ with a \text{Linear} layer. Through the 2d convolution module, potential embeddings of entity $h$ and relation $r$ are jointly encapsulated.

Finally, we compute a similarity score for each triple $(h, r, t)$ with a dot product similarity function:
\begin{equation}
\beta(h,r,t) = \varphi (h,r) . \mathbf{t}
\label{sim}
\end{equation} 
where $\mathbf{t}$ $\in$ $\mathbf{E}$.
When predicting the head $h$ based on pair $(r, t)$, we reverse the relation by adding a special symbol, and obtain a new triple $(t, r_{\text{rev}}, h)$. The similarity score for $(t, r_{\text{rev}}, h)$ is computed in a similar fashion as above following \Cref{bigru}-\Cref{sim}.




\subsection{Contrastive Entity}
\label{subsec:CE}

Contrastive Entity (\Cref{model}B) alleviates sparsity of OpenKGs from the perspective of \emph{nagative} entities, and induces discriminative features of different entities with the same $(h, r)$-pair.
The contrastive score for a triplet $p_e= (h, r, t^{+})$ is:
\begin{equation}
\small 
	S(h,r,t^{+}) =- \log \frac{e^{(\beta(h,r,t^{+})/\tau)}} { \sum_{n \in \{ p_e, \mathcal{N}_e \}} e^{(\beta(n)/\tau)}} 
	\label{entity2}
\end{equation}
where 
$\tau$ is a temperature hyperparameter, $\beta(.)$ is the similarity score as in \Cref{sim}, and $p_e = (h, r, t^{+})$ is a true triple in the OpenKG; $\mathcal{N}_e = \{(h,r,t_j^{-})\}_{j=1}^{|\mathcal{N}_e|}$ is a set of negative samples, where a negative entity $t_j^{-}$ is selected from a candidate entity list defined by: $\mathcal{E}-\mathcal{E}(h,r)$ with $\mathcal{E}(h,r)$ being the entity list of true answers (tail entities), that is, $t_i \in \mathcal{E}(h,r)$ if the triple ($h,r,t_i$) $\in \mathcal{G}$. To analyse how this contrastive loss affects the learning, we perform gradient analysis. It can be shown that the gradients with respect to the head entities are:
\begin{equation}
\small 
\begin{split}
- \frac{\partial S(h,r,t^{+})}{\partial \mathbf{h}} = \frac{\varphi '(h,r) }{\tau A} \Big(  \big[ \hspace{-1em} \sum_{(h,r,t_j^-) \in \mathcal{N}_e} \hspace{-1em} e^{(\frac{  \varphi (h,r) \cdot \mathbf{t_j^-} }{\tau})} \big] \mathbf{t^+} \\
    - \hspace{-1em} {\sum_{(h,r,t_j^-) \in \mathcal{N}_e} } \hspace{-1em} [e^{(\frac{  \varphi (h,r) \cdot \mathbf{t_j^-}  }{\tau})}\mathbf{t_j^-}]
    \Big)
    \label{gradienth}
\end{split}
\normalsize
\end{equation}
where $A$ is a normalization constant (see Appendix for a derivation). This is consistent with our intuition, where positive entity $t^+$ gives positive feedback while negative entities $t_j^-$ give negative feedback. The gradients with respect to the relations ($- \frac{\partial S(h,r,t^{+})}{\partial \mathbf{r}}$) has a similar form as \Cref{gradienth}.

Similarly, we can derive the gradients for the positive $t^+$ and negative tail entities $t_j^{-}$ as:
\begin{equation}
\small 
\begin{split}
	- \frac{\partial S(h,r,t^{+})}{\partial \mathbf{t^+}} 
    &= \frac{\varphi (h,r) }{\tau A} \sum\limits_{(h,r,t_j^-) \in \mathcal{N}_e } e^{(\frac{  \varphi (h,r) \cdot \mathbf{t_j^-} }{\tau})}
    \label{gradienthj+}
\end{split}
\normalsize
\end{equation}
\begin{equation}
\small 
\begin{split}
	- \frac{\partial S(h,r,t^{+})}{\partial \mathbf{t_j^-}}
    &= - \frac{\varphi (h,r)}{\tau A} e^{(\frac{  \varphi (h,r) \cdot \mathbf{t_j^-} }{\tau})}
    \label{gradienthj_}
\end{split}
\normalsize
\end{equation}
For a few-shot entity $t$, when it appears as a positive (\Cref{gradienthj+}) or negative (\Cref{gradienthj_}) sample headed by entity $h$ of high degree, it gets sufficient gradients to learn informative representations. Similarly, through \Cref{gradienthj_}, the parameters of a zero-shot entity get updated when it appears as a negative sample with a $(h, r)$-pair. As discussed later in \cref{subsec:CSF}, with Contrastive Self, zero-shot entities get updated with their own contrastive losses. Overall, through the negative samples, training of few-shot, zero-shot and other entities (with many links) gets more balanced, while with existing approaches, zero-shot entities generally do not get trained as they are disconnected from the rest of the OpenKG.

\subsection{Contrastive Relation}
\label{subsec:CR}

Entities in a KG propagate information to neighboring entities through one or more relations, so the features of relations are as important as that of entities. Contrastive Relation (\Cref{model}C) alleviates the sparsity from the perspective of \emph{negative} relations, and captures potential features of different relations with the same $(h, t)$-pair. The contrastive score for a positive relation $p_r= (h, r^{+}, t)$ is:
\begin{equation}
\small
	S(h,r^{+},t) =- \log \frac{e^{(\beta(h,r^{+},t)/\tau)}}{\sum_{n \in \{p_r, \mathcal{N}_r\} } e^{(\beta(n)/\tau)}} 
	\label{relation2}
\end{equation}
where $p_r$ is a true triple in the OpenKG, and $\mathcal{N}_r = \{(h, r_j^{-}, t)\}_{j=1}^{|\mathcal{N}_r|}$ is a set of negative samples, where a negative relation $r_j^{-}$ is sampled from a candidate relation list $\mathcal{R}-\mathcal{R}(h,t)$ with $\mathcal{R}(h,t)$ being a relation list that satisfies the condition: $r_i \in \mathcal{R}(h,t)$ if the triple ($h,r_i,t$) $\in \mathcal{G}$.
The gradients have the same form as above (see Appendix A.1). In particular, the gradients that a tail entity gets is: 
\begin{equation}
\small 
\begin{aligned}
    - \frac{\partial S(h,r^{+},t)}{\partial \mathbf{t}} = &\frac{1}{\tau B} \Big( \big[ \hspace{-1em} \sum\limits_{(h,r_j^-,t) \in \mathcal{N}_r } \hspace{-1em} e^{(\frac{  \varphi (h,r_j^-) \cdot \mathbf{t} }{\tau})} \big] \varphi (h,r^+) \\
	&- \hspace{-1em} \sum\limits_{(h,r_j^-,t) \in \mathcal{N}_r } \hspace{-1em} [e^{(\frac{  \varphi (h,r_j^-) \cdot \mathbf{t} }{\tau})} \varphi (h,r_j^-)]  
	\Big) 
\end{aligned}
\label{rel_gradientt}
\end{equation}
Different from \Cref{gradienthj+}, tail entities ($t=t^+$) get contrastive gradients, because the signals from the positive ($r^+$) and negative ($r_j^-$) samples are in opposite direction in \Cref{rel_gradientt}. 





	



\subsection{Contrastive Self}
\label{subsec:CSF}

As described in \cref{subsec:CE} and \cref{subsec:CR}, parameters of a few-shot entity get updated with contrastive gradients when it acts as a head (\Cref{gradienth}) or tail entity (\Cref{rel_gradientt}). Parameters of a zero-shot entity are also updated when it appears as a negative entity (\Cref{gradienthj_}), but not from the perspective of contrastive. This means zero-shot entities have no chance to learn discriminative representations, because they have no links with the rest of OpenKG.

In view of this, we propose to construct a positive sample $p_{sf}=(h, r_{sf}, h^{+})$ by adding a \emph{self} relation (\Cref{model}D), where $h$ and $h^{+}$ are the same entity but with different embeddings: the embedding of $h$ is $[\mathbf{h} + \mathbf{h^w}]$ and the embedding of $h^{+}$ is $\mathbf{h} \in \mathbf{E}$. For such a positive sample, negative samples $\mathcal{N}_e = \{(h,r_{sf},h_j^{-})\}_{j=1}^{|\mathcal{N}_e|}$ are generated with negative entities by selecting $h_j^{-}$ strategically from an entity list $\mathcal{E}-h$ (\Cref{model}B-D). Similarly, negative relation samples $\mathcal{N}_r = \{(h,\{r_{sf}\}_j^{-},h^{+})\}_{j=1}^{|\mathcal{N}_r|}$ are generated with negative relations by selecting $\{r_{sf}\}_j^{-}$ strategically from a relation list $\mathcal{R}-r_{sf}$ (\Cref{model}C-D). The contrastive scores for $p_{sf}=(h,r_{sf},h^{+})$ can then be computed with \Cref{entity2} and \Cref{relation2}.

Through the \emph{self} positive sample, parameters of zero-shot entities can have the chances to be updated from the contrastive  perspective (\Cref{gradienth} and \Cref{rel_gradientt}), where $h^+, r_{sf}^+$ give positive feedback while $h_j^-$, $\{r_{sf}\}_j^{-}$ give negative feedback.

\begin{figure}
	\centering
	\includegraphics[width=1.0\columnwidth]{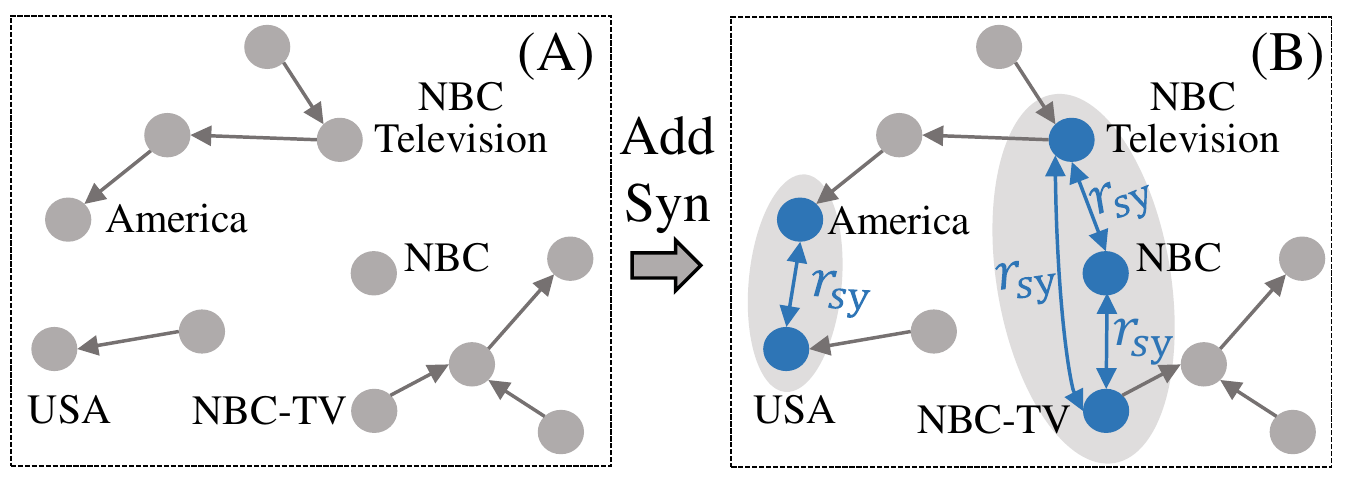}
	\caption{Examples of Contrastive Synonym. (A) A given OpenKG, where two entities \emph{“America”}, \emph{“USA”} are synonymous, three entities \emph{“NBC Television”}, \emph{“NBC”}, \emph{“NBC-TV”} are synonymous. (B) Add a \emph{sy} relation ($r_{sy}$) between each pair of synonyms.
	}
	\label{model_syn}
\end{figure}

\subsection{{Contrastive Synonym}}  
\label{subsec:CSM}

We propose Contrastive Synonym based on the synonymity characteristic of an OpenKG, where multiple entities with different surface forms have the same meaning, e.g., \emph{“NBC-TV”, “NBC”} and \emph{“NBC Television”} mean the same thing. 
These synonymous entities are designed as positive samples of each other.
We identify synonyms automatically by a fuzzy matching with the IDF-token-overlap tool \cite{GalarragaHMS14} and a semantic matching with \Cref{sim} based on the embeddings obtained by \Cref{bigru}. We put the extracted synonymous entities for $h$ into $\mathbb{S}(h)$ and construct a positive sample $p_{sy}=(h, r_{sy}, h_i^{+})$ by adding a \emph{sy} relation for each entity $h_i^{+} \in \mathbb{S}(h)$ (\Cref{model_syn}).
We follow the same strategy as in \Cref{subsec:CE} and \Cref{subsec:CR} for sampling negative entity and relation samples, and compute the contrastive score for each $p_{sy}=(h,r_{sy},h_i^{+})$ with \Cref{entity2} and \Cref{relation2}.



\subsection{Contrastive Fusion}
\label{subsec:CF}

The contrastive learning of the above modules has the form of 1-to-1, where a head entity propagates information to a tail entity through a relation, i.e., it involves only one positive sample. However, a head entity can connect to multiple tail entities through multiple relations. To model this multi-propagation pattern, we extend the 1-to-1 to 1-to-N, where multiple positive samples are considered.

We design two sets of positive 1-to-N patterns: $p_e = \{(h,r,t_j^{+})\}_{j=1}^{|p_e|}$ where a head entity $h$ connects to multiple tail entities through a relation $r$, 
and $p_r = \{(h,r_j^{+},t)\}_{j=1}^{|p_r|}$ where a head entity $h$ connects to a tail entity $t$ through multiple relations. We generate negative samples $\mathcal{N}_e =\{(h,r,t_j^{-})\}_{j=1}^{|\mathcal{N}_e|}$ for $p_e$ as in \cref{subsec:CE}, and negative samples $\mathcal{N}_r =\{(h,r_j^{-},t)\}_{j=1}^{|\mathcal{N}_r|}$ for $p_r$  as in \cref{subsec:CR}. Let $p_f=\{p_e, p_r\}$ and $\mathcal{N}_f=\{\mathcal{N}_e, \mathcal{N}_r\}$. With this, we design two types of contrastive scores to learn 1-to-N patterns.
\begin{equation}
\small 
 S_a(p_f) =- \log \frac{\sum_{n^+ \in p_f } e^{(\beta(n^+)/\tau)}}
	{\sum_{n^- \in \{ p_f,\mathcal{N}_f \} } e^{(\beta(n^-)/\tau)}} \label{fusion1}
\end{equation}
\begin{equation}
\small 
	 S_b(p_f) =- \hspace{-0.3em} \sum_{n^+ \in p_f } \log \frac{e^{(\beta(n^+)/\tau)}}{\sum_{n^- \in \{ n^+,\mathcal{N}_f\} } e^{(\beta(n^-)/\tau)}} 
	\label{fusion4}
\end{equation}
In \Cref{fusion1}, all positive samples are trained together and normalized  to a unified space, 
while in \Cref{fusion4}, different positive samples are trained separately and the independent similarity score of all positives are accumulated. 
From the perspective of negative samples, negative entity $\mathcal{N}_e$ and relation $\mathcal{N}_r$ samples are merged to train with the positive samples, which is different from \Cref{entity2} and \Cref{relation2} which only have one type of negatives.

\begin{table*}[t]
	\centering
	\resizebox{1.0\textwidth}{!}{
		\begin{tabular}{ccccccccccccccc}
             \toprule
			 \multirow{2}{*}[-1pt]{Dataset} & \multirow{2}{*}[-1pt]{Ent} & \multirow{2}{*}[-1pt]{Rel} & \multirow{2}{*}[-1pt]{Clust}  & 
			 \multirow{2}{*}[-1pt]{Valid}  & 
			 \multirow{2}{*}[-1pt]{Test}   & 
			 \multicolumn{5}{c}{Train (Sparsity Levels)} & \multicolumn{2}{c}{Few-Shot} & \multicolumn{2}{c}{Zero-Shot}  \\
			 \cmidrule(lr){7-11}\cmidrule(lr){12-13} \cmidrule(lr){14-15}
			 & & & & & & 100\% & 80\% & 60\% & 40\% & 20\% & Ent & Rel & Ent & Rel \\
             \midrule
			\textbf{ReVerb20K}    & 11.1   & 11.1 & 10.8       & 1.6   & 2.4 & 15.5 & 12.4 & 9.3 & 6.2 & 3.1 & 2.8 / 1.0 & 2.5 / 0.8 & 8.1 / 1.1 & 8.5 / 1.0 \\
			\textbf{ReVerb45K}    & 27.0   & 21.6 & 18.6       & 3.6   & 5.4 & 36.0 & 28.8 & 21.6 & 14.4 & 7.2 & 7.7 / 1.8 & 5.1 / 1.5  & 18.8 / 2.5 & 16.4 / 1.8  \\	
 			\bottomrule
		\end{tabular}
	}
	\caption{Dataset statistics (in 1000s). 
	Ent, Rel, Clust show the no. of entities, relations, entity clusters. Valid, Test, Train show the no. of triples in valid, test, train sets. In Train, $x\%$ represents sparsity levels. 
	In Few- and Zero-Shot, left of / respresents the no. of entities or relations, and right of / represents the no. of related triples in its Test set.}
	\label{datasets}	
\end{table*}

\subsection{Training Procedure} 
\label{subsec:train}
We train TernaryCL in Pretrain and Finetune stages, where Pretrain stage aims to learn discriminative representations with \Cref{fusion1} or \Cref{fusion4}, and Finetune stage aims to optimize parameters for a target task. 
Note that TernaryCL is a general framework that can be applied to diverse KG types, such as OpenKGs, CuratedKGs and TemporalKGs, and to diverse applications, such as link prediction, relation prediction, node classification, relation extraction and other ternary tasks. In addition, the principle of Contrastive Self can be applied to dynamic tasks, e.g., when a new entity gets added to a KG, it can transfer embeddings of existing entities to the new entity. Since our focus in this work is the sparsity of OpenKGs, we design experiments with link prediction tasks in OpenKGs.

For a test triple $(h_i,r_i,t_i)$, the jointly encapsulated representation $\varphi (h_i, r_i)$ of entity $h_i$ and relation $r_i$ is matched with the embeddings of all the entities in $\mathbf{E}$ to predict $\hat{Y}_i \in [0,1]^{|\mathcal{E}|}$ as:
\begin{equation}
\small 
	\hat{Y}_i = \text{sigmoid} (\varphi (h_i, r_i) \cdot \mathbf{E}^{\top})
	\label{trainingp1}
\end{equation}
We use a binary cross-entropy loss defined as:
\begin{equation}
\small 
- \frac{1}{|\mathcal{E}|} \sum_{j=1}^{|\mathcal{E}|} Y_{i,j}  \log \hat{Y}_{i,j} + (1-Y_{i,j}) \log (1-\hat{Y}_{i,j})
\label{trainingp2}
\end{equation}
where $Y_{i,j} = 1 \text{ if} \ t_j \in \mathcal{E}(h_i,r_i)$ otherwise $Y_{i,j} = 0$ with  $\mathcal{E}(h,r)$ being the set of all true tail entities for a pair $(h,r)$. As mentioned earlier, when predicting the head $h_i$ based on a pair $(r_i, t_i)$, we reverse the relation to obtain a reversed triple $(t, r_{\text{rev}}, h)$, then train it with \Cref{trainingp1} and \Cref{trainingp2}.

\section{Experiments} \label{sec:exp}

\subsection{Datasets and Experiment Setup}

\label{sec:expdatasets}

We use ReVerb20K and ReVerb45K OpenKG benchmarks \cite{VashishthJT18}, which are constructed through ReVerb \cite{FaderSE11}. Table \ref{datasets} presents their statistics. 
ReVerb45K with 27K entities and 21.6K relations is larger and sparser than ReVerb20K with 11.1K entities and 11.1K relations. 
Entity clusters are gold canonicalized clusters, extracted through the Freebase entity linking tools \cite{GuptaKT19}.
Entities in an entity cluster have the same meaning. 
Usually, entity clusters are only used for evaluation.


To evaluate on sparsity, we design two sets of controlled experiments: the first is at different sparsity levels and the second is on few- and zero-shot samples. For the first, 
we construct \emph{train sets} at different sparsity granularity \{100\%, 80\%, 60\%, 40\%, 20\%\} by respectively removing \{0\%, 20\%, 40\%, 60\%, 80\%\} of the links from the original train set. We use the same original data for validation and testing. For the second, we evaluate on few-shot or zero-shot samples separately, where few-shot refers to 3-, 2- and 1-shot. Few- or zero-shot entities (relations) are extracted as per the definition in \cref{preliminaries}, e.g., a 3-shot entity has three links in the test KG. 
For the test set of a few-shot (zero-shot) entity (relation), its related triples are extracted from the original test set.
For training, we use the 20\% train set, because it is sparse enough to contain more few-shot (zero-shot) samples. 



\subsection{Evaluation Metrics and Baselines}

To evaluate on a single test triple, we use Mention Ranking or MR \cite{GuptaKT19}, which is the minimum ranking position of the answer entities. To evaluate over all test triples, we use the three most widely used ways to integrate the individual MR scores: (a) $H@N$: proportion of MR scores not higher than $N$, (b) $AR$: average of all MR scores, and (c) $ARR$: compute the reciprocal of each MR score, and average all reciprocals. 
A model with better performance should have higher $H@N$ and $ARR$ scores and a lower $AR$ score.\footnote{In previous work, $ARR$ and $AR$ are called $MRR$, and $MR$ respectively, where $M$ stands for \emph{Mean}. However, since \emph{Mention Ranking} is abbreviated as $MR$, to prevent confusion, we use \emph{Average} in the names in stead of \emph{Mean}.}

\begin{table*}[t]
	\centering
	\resizebox{0.98\textwidth}{!}{
	\begin{tabular}{clcccccccccccc}
		\toprule
		\multirow{2}{*}[-2pt]{Type} & \multirow{2}{*}[-2pt]{Model} & \multicolumn{6}{c}{ReVerb20K} & \multicolumn{6}{c}{ReVerb45K} \\
        \cmidrule(lr){3-8}\cmidrule(lr){9-14}
		& & $AR \downarrow$ & $ARR$ & $H@$1 & $H@$10 & $H@$50 & $H@$100
		& $AR \downarrow$ & $ARR$ & $H@$1 & $H@$10 & $H@$50 & $H@$100 \\
		\midrule
		\multirow{5}{*}{General}
		& TransE 	 & 1497   & 13.3 & 2.2 & 29.6 & 43.0 & 49.2 & 2222 & 15.8  & 9.3  & 25.9 & 37.1 & 43.2 \\    
		& DistMult    
		& 4569   & 1.9  & 1.3 & 2.7  & 5.2 & 7.1 
		& 5782 & 8.5   & 7.7  & 9.7  & 12.0 & 13.6 \\
		& ComlEx     
		& 4376     & 2.0  & 1.4 & 3.0   & 5.6 & 7.7 
		& 5173 & 8.9  & 7.5  & 11.3  & 16.0 & 18.9 \\
		& ConvE 	     
		& 1085     & 25.5 & 19.9 & 35.8 & 50.1 & 57.2 
		& 2483 & 22.1 & 16.6  & 32.4 & 43.3 & 47.9 \\
		& ConvTransE 	     
		& 1080 & 26.1 & 20.5 & 35.9 & 50.0 & 57.1 
		& 2490 & 23.4 & 17.9  & 33.8 & 44.4 & 48.8 \\
		
         \midrule
		\multirow{2}{*}{OpenKG}
		& CaReTransE    
		& 950  & 30.3 & 23.2 & 42.8 & 58.4 & 64.6 
		& 2414 & 19.5 & 7.8 & 37.5 & 47.5 & 51.4 \\
	    & CaReConvE 	
		& 801  & 31.6 & 25.6 & 42.9 & 56.7 & 63.4  
		& 1589 & 29.7 & 23.4 & 41.3 & 53.6 & 58.7 \\
		

        \midrule
        \multirow{2}{*}{OpenKG}
        & SimKGC 
		& 538  & 29.7 & 23.3 & 41.7 & 58.7 & 65.7 
		& 1080 & 27.1 & 20.4 & 39.8 & 53.2 & 59.2 \\
		\multirow{2}{*}{+PLM} & OKGITBert 
		& 524 & 35.1 & 27.5 & 49.5 & 65.9 & 72.7 
		& \textbf{735} & \textbf{33.7} & \textbf{26.7} & 47.1 & 59.8 & 65.2 \\
		& OKGITRob 
		& 594 & 35.8 & 28.4 & 49.2 & 65.4 & 72.1
		& 849 & 33.4 & 26.5 & 46.4 & 58.8 & 63.9 \\
 		\midrule
		\textbf{Our} & \textbf{TernaryCL} 
		& \textbf{393} & \textbf{38.9} & \textbf{30.5} & \textbf{54.6} & \textbf{69.2} & \textbf{75.5} 
		& 767 & 33.3 & 25.3 & \textbf{48.7} & \textbf{63.0} & \textbf{68.3} \\
 		\bottomrule
	\end{tabular}
	}
	\caption{Results on ReVerb20K and ReVerb45K in the standard full data (100\% train) setup. Best scores are made bold. Columns with $\downarrow$ denote lower is better, otherwise higher is better. }
	\label{rv_45}
\end{table*}

\begin{figure*}[t]
    \begin{minipage}[t]{0.38\textwidth}
    \centering
    \setlength{\abovecaptionskip}{0cm} 
    \setlength{\belowcaptionskip}{0mm}
    \subfigure{
    \includegraphics[width=1\columnwidth]{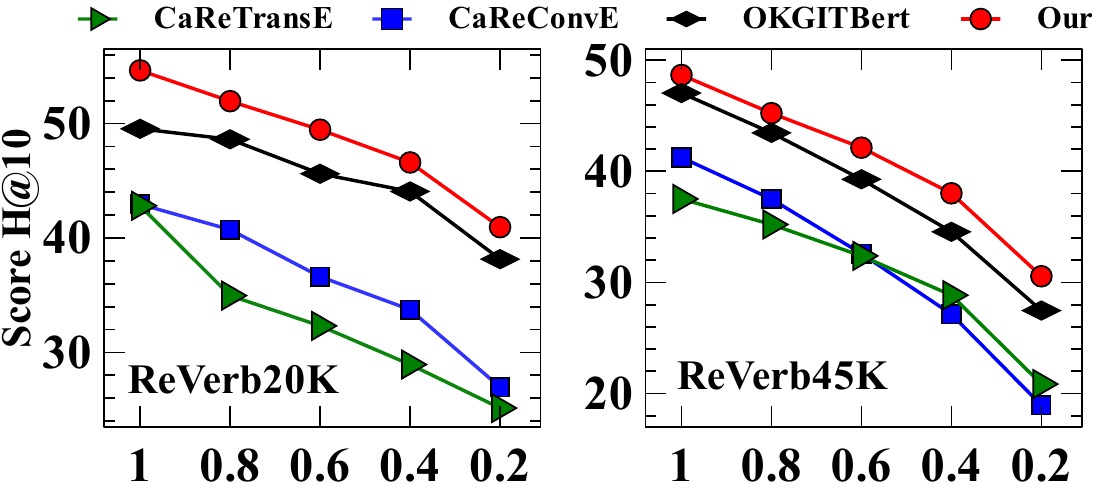}
    }
    \caption{Results at sparsity levels on ReVerb20K (Left) and ReVerb45K (Right).}
	\label{sparsity}
    \end{minipage}
    \hspace{2mm}
    \begin{minipage}[t]{0.6\textwidth}
    \centering
    \setlength{\abovecaptionskip}{0cm} 
    \setlength{\belowcaptionskip}{0mm}
    \subfigure[\small rv20-B]{
    \includegraphics[width=0.22\columnwidth]{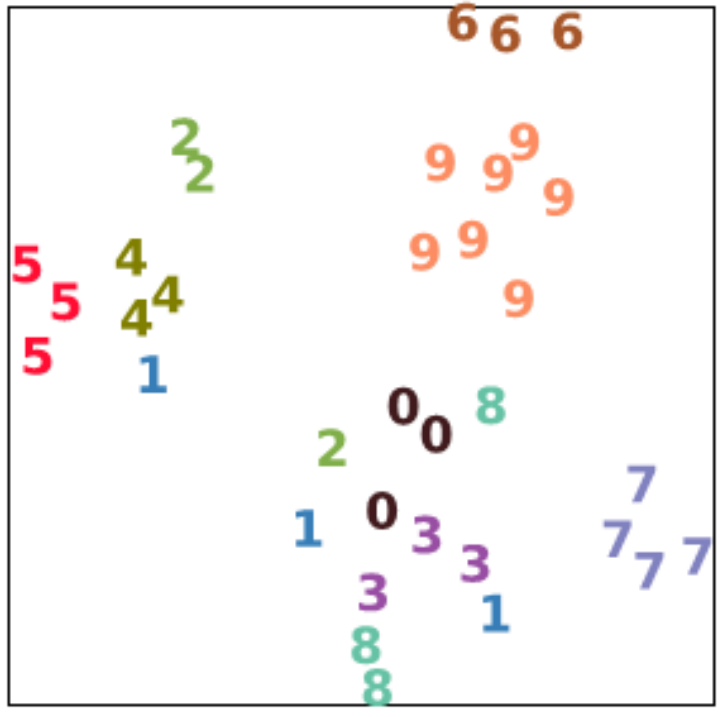}
    }
    \subfigure[\small rv20-O]{
    \includegraphics[width=0.22\columnwidth]{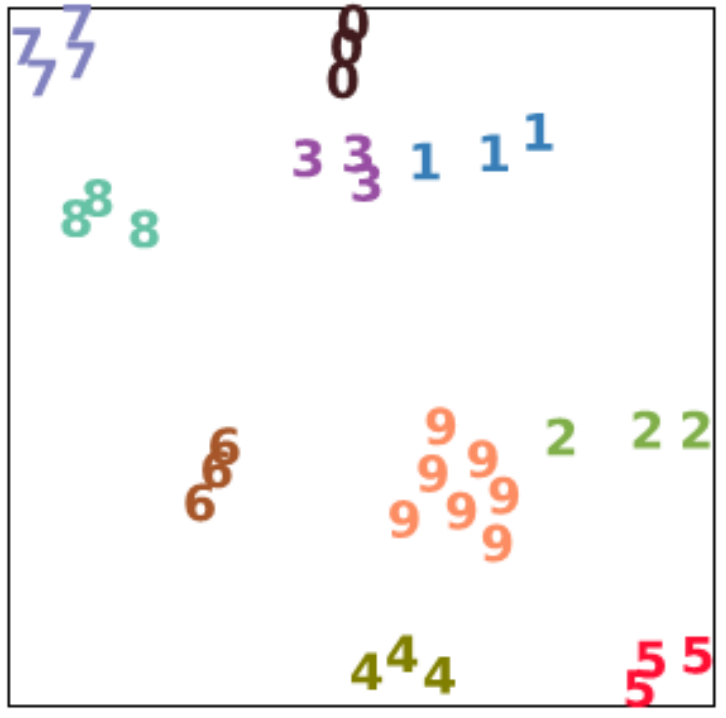}
    }
    \subfigure[\small rv45-B]{
    \includegraphics[width=0.22\columnwidth]{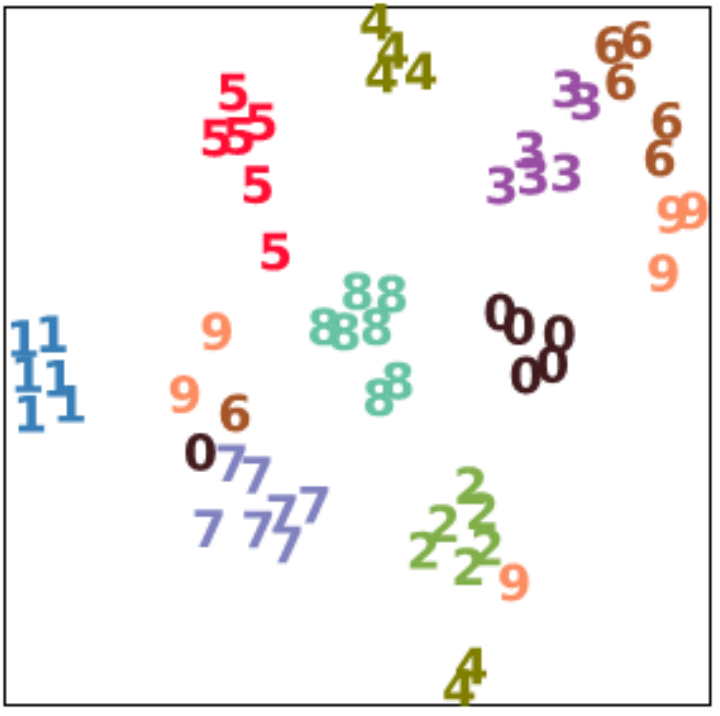}
    }
    \subfigure[\small rv45-O]{
    \includegraphics[width=0.22\columnwidth]{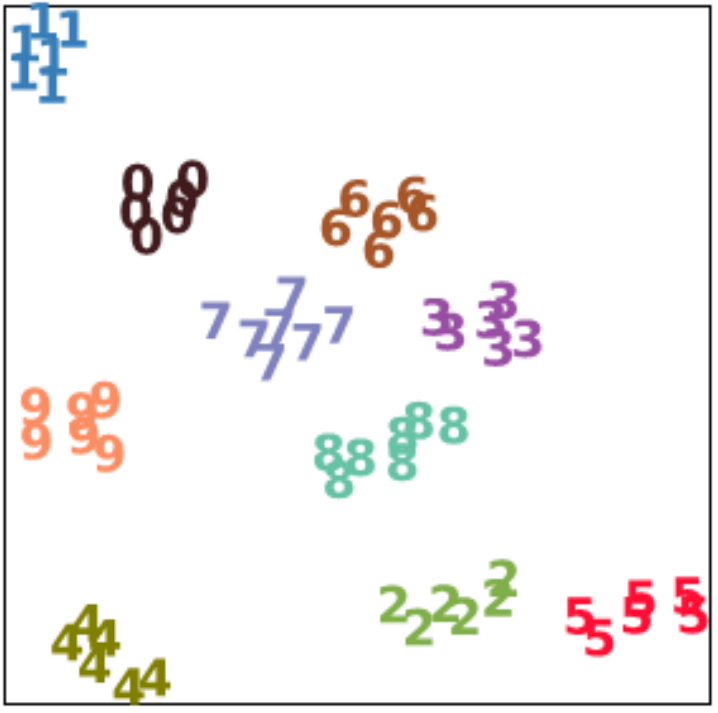}
    }
    \caption{Visualization of Baseline(B) and Our(O) on ReVerb20K (rv20) and ReVerb45K(rv45). Same numbers denote similar entities.}
    \label{visual}
    \end{minipage}
\end{figure*}

To show the effectiveness of our approach, we compare TernaryCL against several strong baselines which fall in three groups:

\begin{itemize}[leftmargin=*]
    \item \textbf{General}: This applies general KG models to OpenKGs. It includes TransE \cite{bordes2013translating}, DistMult \cite{yang2014embedding}, ComplEx \cite{trouillon2016complex}, ConvE \cite{dettmers2018convolutional} and ConvTransE \cite{ShangTHBHZ19}.
    \item \textbf{OpenKG}: This comprises the baselines designed specifically for OpenKGs. It includes CaReTransE and CaReConvE \cite{GuptaKT19}.
    \item \textbf{OpenKG+PLM}: This group uses pretrained language models (PLM) to improve the representation learning. It includes SimKGC \cite{0046ZWL22}, OKGITBert and OKGITRob \cite{ChandrahasT21}.
    SimKGC is a state-of-the-art contrastive model for CuratedKGs, we apply it to OpenKGs.
    OKGITBert and OKGITRob are state-of-the-art models for OpenKGs.
\end{itemize}



\paragraph{For our TernaryCL}
TernaryCL is implemented based on the PyTorch library with a single GeForce RTX 2080 GPU.
The number of parameters is 18.06M for ReVerb20K and 33.63M for ReVerb45K. We run the model five times and report the maximum of results.
For the training settings, the optimizer is set to Adam, the embedding size is set to 300. The entity and relation embeddings are initialized randomly, and the word vectors are initialized with the GloVe embeddings. Default fusion strategy (\cref{subsec:CF}) is $S_b(p_f)$ and text encoder (\Cref{subsec:emb}) is BiGRU. 

We tune our model with the grid search to select the optimal hyper-parameters based on the performance on the validation dataset.
The list of hyperparameters are from two aspects: 
(1) Hyperparameters for Finetune stage:
batch size $\in$ \{32, 64, 128, 256, 512\}, learning rate $\in$ \{1e-3, 1e-4, 8e-5, 5e-5, 1e-5\}.
(2) Hyperparameters for Pretrain stage:
learning rate $\in$ \{1e-3, 1e-4, 5e-5, 1e-5\}, temperature regulation value $\in$ \{0.1, 0.05, 0.01\}.
The results of hyperparameters are shown in \Cref{all}. 

\begin{table*}[t]
	\centering
	\resizebox{1.0\textwidth}{!}{
	\begin{tabular}{l cccccccc cccccccc}
		\toprule
		\multirow{2}{*}[-5pt]{Model} &
        \multicolumn{4}{c}{Few-Shot Entity} &
		\multicolumn{4}{c}{Few-Shot Relation} &
		\multicolumn{4}{c}{Zero-Shot Entity} &
		\multicolumn{4}{c}{Zero-Shot Relation} \\
        \cmidrule(lr){2-5}
        \cmidrule(lr){6-9}
        \cmidrule(lr){10-13}
        \cmidrule(lr){14-17}
	    & \small $AR \downarrow$ & \small $ARR$ & \small $H@$1 & \small $H@$10 
	    & \small $AR \downarrow$ & \small $ARR$ & \small $H@$1 & \small $H@$10 
	    & \small $AR \downarrow$ & \small $ARR$ & \small $H@$1 & \small $H@$10
	    & \small $AR \downarrow$ & \small $ARR$ & \small $H@$1 & \small $H@$10 \\
		\midrule
		\small CaReTransE & 1215 & 10.6 & 1.2  & 25.6   & 2022 & 10.7  & 1.7 & 26.4
		           & 3286 & 7.3 & 0.0   & 20.9   & 2071 & 10.4  & 2.2 & 24.8 \\
		\small CaReConvE  & 2196 & 24.4 & 19.6 & 33.1   & 2663 & 20.6  & 16.2 & 29.3
		           & 3304 & 14.2 & 11.1 & 20.8   & 2557 & 22.3  & 18.1 & 30.4 \\
		\small OKGITBert  & \textbf{285} & 29.4 & 22.2  & 43.1   & 1696 & 26.2  & 19.3 & 38.8
		           & 3465 & 20.1 & 15.1 & 29.8   & 1775 & 27.4  & 20.7 & 40.8 \\
		\small TernaryCL & 364 & \textbf{33.1} &	\textbf{25.0} & \textbf{48.6} & \textbf{863} & \textbf{27.7} & \textbf{20.5} & \textbf{41.8} 
		& \textbf{1473} & \textbf{20.7} & \textbf{15.5} & \textbf{31.1} & \textbf{868} & \textbf{30.1} & \textbf{23.1} & \textbf{43.2} \\
		
		\midrule
		\small CaReTransE & 2610 & 12.6 & 6.7  & 21.4    & 3323 & 13.6  & 7.8 & 23.0
		           & 5776 & 10.2 & 8.8  & 12.4    & 3368 & 11.8  & 6.4 & 20.3 \\
		\small CaReConvE  & 2739 & 19.2 & 15.5 & 26.2    & 3379 & 14.8  & 10.9 & 22.8
		           & 5662 & 4.3 & 3.0   & 6.6     & 3429 & 12.5  & 8.8 & 19.2 \\
		\small OKGITBert  & \textbf{644} & 23.5 & \textbf{18.2}  & 33.7    & 2211 & 20.0  & 14.5 & 31.0
		           & 4952 & 7.9 & 5.4   & 12.5    & 2349 & \textbf{18.6}  & \textbf{13.2} & 29.2 \\
		\small TernaryCL & 954 & \textbf{23.6} &	16.9 & \textbf{35.8} & \textbf{1460} & \textbf{21.3} & \textbf{14.9} & \textbf{32.8} & \textbf{2626} &	\textbf{13.2} & \textbf{9.9} & \textbf{19.5} & \textbf{1539} & 18.3 & 12.3 &	\textbf{29.9} \\
		
		\bottomrule
	\end{tabular}
	}
	\caption{Results of few-/zero-shot entities/relations on ReVerb20K (Top) and ReVerb45K (Below) with 20\% train.
	}
	\label{fewshot}
\end{table*}

\paragraph{For Baselines}
The results of baselines are reproduced with open source implementations.
Concretely, TransE, DistMult and ComplEx are reproduced with public code in \footnote{https://github.com/uma-pi1/kge \label{web}}. The code for ConvE is in \footnote{ https://github.com/malllabiisc/CaRE \label{webcare}} and for SimKGC is in \footnote{ https://github.com/intfloat/SimKGC \label{SimKGC}}.
The code for ConvTransE is implemented by us and public in our code.
We use the grid search technique to select the optimal values of hyperparameters for above baselines. 
For OpenKG and OpenKG+PLM baselines, CaReTransE, CaReConvE are reproduced with the public code in \textsuperscript{\ref{webcare}}, OKGITBert and OKGIT+Rob are reproduced with the public code in \footnote{https://github.com/Chandrahasd/OKGIT}. 
The optimal values of hyperparameters for this four baselines are consistent with that in their paper.

\subsection{Results}
\label{results1}

\paragraph{$\bullet$ Full-data evaluation.} We first present the results on the standard full-data setup in Table \ref{rv_45}, for which we use the original train set (100\%). We notice that TernaryCL achieves substantial improvements in comparison to the baselines.
For ReVerb20K, it outperforms all the baselines by a good margin across all the metrics -- 
ARR increases by 3.1 point and $H@$1,10,50,100 increase by 2.1, 5.1, 3.3, 2.8 points, respectively.
For ReVerb45K, its performance is better than General and OpenKG baselines with sizeable margins in all metrics, notably 3.6 point in ARR and 1.9, 7.4, 9.4, 9.6 points in $H@$1,10,50,100. It also outperforms OpenKG+PLM baselines in all metrics for ReVerb20K and in 3 of 6 metrics for ReVerb45K.

Overall, TernaryCL with a simple structure can achieve better performance through innovative training methods at lower costs than OpenKG+PLM baselines that use a relatively complex structure and large PLMs which is memory and compute intensive. This can make TernaryCL a favorable choice against the baselines.


\paragraph{$\bullet$ Targeted evaluation with sparsity.}
We now evaluate whether TernaryCL can alleviate the sparsity problem from the perspective of different sparsity levels. \Cref{sparsity} reports the results on the (original) test set for different train sets with varying sparsity level. We observe that TernaryCL achieves the best scores at all sparsity levels on both datasets. Taking ReVerb20K as an example, it gives improvements of 3.3, 3.8, 2.5 and 2.8 points over OKGITBert at 80\%, 60\%, 40\%, 20\%, respectively. Note that OKGITBert enhances learning with large-scale PLMs, which possess a lot of commonsense and factual knowledge within its huge parameter space by training on large data. In contrast, TernaryCL does not use side information or PLMs, but gives significant performance gains over OKGITBert, even when the sparsity level is high (e.g., 20\%).

\paragraph{$\bullet$ Targeted evaluation for few- and zero-shot.}
We now analyze whether TernaryCL can alleviate sparsity on test sets with few- or zero-shot entities (or relations). As shown in \Cref{fewshot}, CaReTransE performs poorly, especially, its $H@$1 score is 0.0 for zero-shot entities on ReVerb20K.
OKGITBert achieves better results than CaReTransE and CaReConvE on most metrics, which shows that pretrained knowledge introduced by Bert could be helpful to few- and zero-shot entities and relations. The proposed TernaryCL, without using any PLM or extra information, outperforms all baselines by a good margin on most metrics. 






\paragraph{$\bullet$ Visualization evaluation.}

To qualitatively show that TernaryCL can make entities with the same meaning closer in the vector space, we show t-SNE visualization \cite{van2008visualizing} in \Cref{visual}. For this, we train TernaryCL on the original train set (100\%) and compare with the state-of-the-art baseline OKGITBert. For visualization, we selected 10 entity clusters from the test set, where each cluster has more than three entities. We observe that entities with the same meaning (same number) are closer in TernaryCL than in OKGITBert. These qualitative results are consistent with the above quantitative results, which further verify the effectiveness of the proposed TernaryCL.

\subsection{Ablation Study and Analysis}

\begin{figure}
	\centering
	\includegraphics[width=1.0\columnwidth]{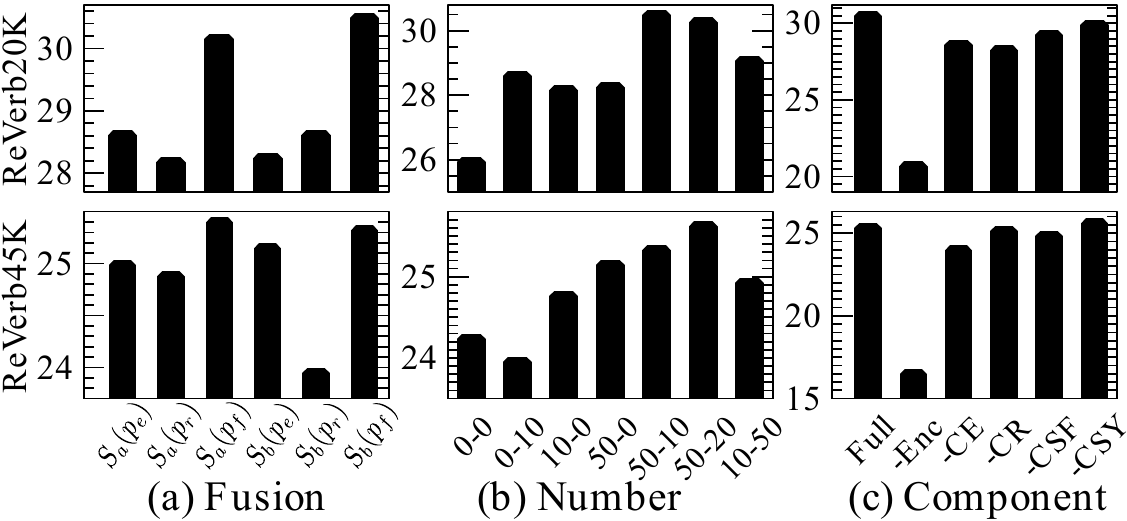}
	\caption{Results ($H@$1) for {(a) fusion strategies}, (b) no. of negative entities and relations, and (c) components}
	\label{ab}
\end{figure}

\paragraph{$\bullet$ Fusion strategy.}
The ablations for the fusion strategies (\Cref{subsec:CF}) are shown in \Cref{ab}(a), where $S_a(p_e)$, $S_a(p_r)$, $S_b(p_e)$, $S_b(p_r)$ use only one type of 1-to-N pattern (denoted by the subscript of $p$), and $S_a(p_f)$ and $S_b(p_f)$ are our two variants. $S_a(p_f)$, which fuses both 1-to-N types, outperforms its counterparts $S_a(p_e)$ and $S_a(p_r)$. We see the same phenomenon with $S_b(p_f)$, which outperforms $S_b(p_e)$ and $S_b(p_r)$. This proves that both 1-to-N types carry different semantic features, and integration of them yields additive gains. When we compare our two variants, $S_b(p_f)$ shows better performance than $S_a(p_f)$ on ReVerb20K while similar performance on ReVerb45K. Beside $H@$1, we also compare the $H@$10,50,100 scores on ReVerb45K; these scores of $S_a(p_f)$ are 48.3, 62.5, 67.9, which are lower than that of $S_b(p_f)$ in \Cref{rv_45}. $S_b(p_f)$ contrasts a positive sample with both negative entities and relations, which is able to capture more discriminative features for the positive sample.







\paragraph{$\bullet$ No. of negative entities and relations.}
To investigae how the number of negative entities and relations could affect the performance of TernaryCL,  we design several collocations of negative entities and negative relations (entity-relation):
0-0, 0-10, 10-0, 50-0, 50-10, 50-20, 10-50. Results in \Cref{ab}(b) show that TernaryCL achieves the best results on ReVerb20K at 50-10 and on ReVerb20K at 50-20 which have the bigger no. of negative entities and the smaller no. of negative relations.
It also reveals that negative entities are more important than negative relations in improving performance.

\paragraph{$\bullet$ Component.}
We now probe the role of each component in our model (\Cref{ab}(c)). -Enc denotes removing the text encoder (\Cref{subsec:emb}) from the full model. Similarly, -CE, -CR, -CSF, and -CSY represent removing the Contrastive Entity (\Cref{subsec:CE}), Contrastive Relation (\Cref{subsec:CR}), Contrastive Self  (\Cref{subsec:CSF}), Contrastive Synonym (\Cref{subsec:CSM}), respectively. Compared to the full model, performance of -Enc decreases prominently meaning that the text encoder plays a crucial role in improving performance. The performance degradation for -CE, -CR, -CSF proves that Contrastive Entity, Contrastive Relation, Contrastive Self are also important components of the model.  
Removing Contrastive Synonym (-CSY) decreases the performance on ReVerb20K while has similar performance on ReVerb45K. Beside $H@$1, we also computed $H@$10,50,100 scores on ReVerb45K; -CSY gets 48.1, 61.9, 67.7, which are lower than ones of the full model (\Cref{rv_45}). 

Note that the roles of BiGRU (-Enc) and CL (-CE, -CR, -CSF, -CSY) are different --  BiGRU is a basic component to encode textual features, while CL is a training mechanism to capture discriminative features. The main take home messages from our experiments and ablations are: (1) as an encoder, Bi-GRU is more effective than BERT; (2) CL gives further significant improvements irrespective of the encoder.

\section{Conclusion}
In this work, we have provided empirical insights about the sparsity problem of OpenKGs, and proposed TernaryCL, a contrastive learning framework to alleviate the sparsity by introducing contrastive entity, relation, self, synonym and fusion methods. Through extensive experiments and comprehensive analysis, we have shown that TernaryCL outperforms state-of-the-art baselines by a good margin.
 

\section*{Limitations}
Text encoder is an important component in the proposed TernaryCL model, where the default is BiGRU. Pretrained language models (PLMs) have achieved great results on most NLP tasks, so we also attempted to use the PLM BERT instead of BiGRU. However, performance of the model with BERT (both tuned and not tuned) as an encoder is weaker than the one with BiGRU. The results are shown in Appendix (A.3). This shows that usage of PLMs may not benefit TernaryCL further. In future, we would like to explore other large-scale PLMs like RoBERTa \cite{liu2019roberta} and ELECTRA \cite{ClarkLLM20}.

\section*{Acknowledgements}
The work was supported by National Natural Science Foundation of China (62172086, 61872074, 62272092) and Chinese Scholarship Council.

\bibliography{anthology,emnlp2022}
\bibliographystyle{acl_natbib}

\appendix
\newpage


\section{Appendix}
\label{sec:appendix}

\subsection{Gradient Update}
\paragraph{Gradients of Contrastive Entity}
Taking the \Cref{entity2} as an example, we give the detailed reasoning steps of gradient about $h$, $r$, $t^+$, $t^{-}$.
\begin{equation}
\small
\begin{array}{l}
	-\frac{\partial S(h,r,t^{+})}{\partial h}
	= \frac{\partial}{\partial h}( \log \frac{e^{(\beta(h,r,t^{+})/\tau)}}{\sum_{n \in \{p_e, \mathcal{N}_e\} } e^{(\beta(n)/\tau)}} ) \\
	= \frac{\partial}{\partial h}(  \frac{\beta(h,r,t^{+})}{\tau}- \log \sum_{n \in \{p_e, \mathcal{N}_e\} } e^{(\frac{\beta(n)}{\tau})} ) \\
	= \frac{\partial}{\partial h} (  \frac{\beta(h,r,t^{+})}{\tau}- 
	\log (e^{(\frac{\beta(h,r,t^{+})}{\tau})} \\ 
	 \ \ \ \ +\sum_{(h,r,t_j^{-}) \in \mathcal{N}_e } e^{(\frac{\beta(h,r,t_j^{-})}{\tau})} ) ) \\
	
	= \frac{\partial}{\partial h} (  \frac{\varphi (h,r) \cdot \mathbf{t^+}}{\tau}- 
	\log (e^{(\frac{\varphi (h,r) \cdot \mathbf{t^+}}{\tau})} \\ 
	 \ \ \ \ +\sum_{(h,r,t_j^-) \in \mathcal{N}_e } e^{(\frac{  \varphi (h,r) \cdot \mathbf{t_j^-} }{\tau})} ) ) \\
	 
	 =  \frac{\varphi '(h,r) }{\tau} \mathbf{t^+} - \\
	\ \ \ \ \frac{e^{(\frac{  \varphi (h,r) \cdot \mathbf{t^+} }{\tau})} \frac{\varphi '(h,r)}{\tau}\mathbf{t^+} + \sum\limits_{(h,r,t_j^-) \in \mathcal{N}_e } e^{(\frac{  \varphi (h,r) \cdot \mathbf{t_j^-} }{\tau})} \frac{\varphi '(h,r) }{\tau} \mathbf{t_j^-} }{e^{(\frac{\varphi (h,r) \cdot \mathbf{t^+}}{\tau})} 
    +\sum\limits_{(h,r,t_j^-) \in \mathcal{N}_e } e^{(\frac{  \varphi (h,r) \cdot \mathbf{t_j^-} }{\tau})}} \\
    
    = \frac{\varphi '(h,r) }{\tau} (\frac{\sum\limits_{(h,r,t_j^-) \in \mathcal{N}_e } e^{(\frac{  \varphi (h,r) \cdot \mathbf{t_j^-} }{\tau})}}{e^{(\frac{\varphi (h,r) \cdot \mathbf{t^+}}{\tau})} 
    +\sum\limits_{(h,r,t_j^-) \in \mathcal{N}_e } e^{(\frac{  \varphi (h,r) \cdot \mathbf{t_j^-} }{\tau})}} \mathbf{t^+} \\
    \ \ \ \ -\frac{\sum\limits_{(h,r,t_j^-) \in \mathcal{N}_e } e^{(\frac{  \varphi (h,r) \cdot \mathbf{t_j^-}  }{\tau})}\mathbf{t_j^-}}{e^{(\frac{\varphi (h,r) \cdot \mathbf{t^+}}{\tau})} 
    +\sum\limits_{(h,r,t_j^-) \in \mathcal{N}_e } e^{(\frac{  \varphi (h,r) \cdot \mathbf{t_j^-} }{\tau})}}
    ) \\
    
    = \frac{\varphi '(h,r) }{\tau A} (\sum\limits_{(h,r,t_j^-) \in \mathcal{N}_e } e^{(\frac{  \varphi (h,r) \cdot \mathbf{t_j^-} }{\tau})} \mathbf{t^+} \\
    \ \ \ \ -\sum\limits_{(h,r,t_j^-) \in \mathcal{N}_e } e^{(\frac{  \varphi (h,r) \cdot \mathbf{t_j^-}  }{\tau})}\mathbf{t_j^-}
    ) \\
    
\end{array}
\label{appendix_gradienth}
\end{equation} 

\begin{equation}
\small
\begin{array}{l}
A = e^{(\frac{\varphi (h,r) \cdot \mathbf{t^+}}{\tau})} 
    +\sum\limits_{(h,r,t_j^-) \in \mathcal{N}_e } e^{(\frac{  \varphi (h,r) \cdot \mathbf{t_j^-} }{\tau})} \\
\end{array}
\label{appendix_A}
\end{equation} 
Parameters of entity $h$ can be updated from contrastive with the gradient in \Cref{appendix_gradienth}:
\begin{equation}
\small
\begin{array}{l}
    \mathbf{h}^{l+1} = \mathbf{h}^{l} - \eta(\frac{\partial S(h,r,t^{+})}{\partial h}) \\
    
    = \mathbf{h}^{l} + \frac{\varphi '(h,r)\eta }{\tau A} \sum\limits_{(h,r,t_j^-) \in \mathcal{N}_e } e^{(\frac{  \varphi (h,r) \cdot \mathbf{t_j^-} }{\tau})} \mathbf{t^+} \\
    \ \ \ \ -\frac{\varphi '(h,r) \eta }{\tau A} \sum\limits_{(h,r,t_j^-) \in \mathcal{N}_e } e^{(\frac{  \varphi (h,r) \cdot \mathbf{t_j^-}  }{\tau})}\mathbf{t_j^-}
\end{array}
\label{gradient_update}
\end{equation}

\begin{equation}
\small
\begin{array}{lr}
\frac{\partial S(h,r,t^{+})}{\partial r} = \frac{\partial S(h,r,t^{+})}{\partial h}
\end{array}
\end{equation}

\begin{equation}
\small
\begin{array}{l}
	-\frac{\partial S(h,r,t^{+})}{\partial t^+}
	= \frac{\partial}{\partial t^+} (  \frac{\varphi (h,r) \cdot \mathbf{t^+}}{\tau}- 
	\log (e^{(\frac{\varphi (h,r) \cdot \mathbf{t^+}}{\tau})} \\ 
	 \ \ \ \  \ \ \ \  \ \ \ \  \ \ \ \  \ \ \ \  \ \ \ \ +\sum_{(h,r,t_j^-) \in \mathcal{N}_e } e^{(\frac{  \varphi (h,r) \cdot \mathbf{t_j^-} }{\tau})} ) ) \\
	 
	 =  \frac{\varphi (h,r) }{\tau} - \\
	\ \ \ \ \frac{e^{(\frac{  \varphi (h,r) \cdot \mathbf{t^+} }{\tau})} \frac{\varphi (h,r)}{\tau}}{e^{(\frac{\varphi (h,r) \cdot \mathbf{t^+}}{\tau})} 
    +\sum\limits_{(h,r,t_j^-) \in \mathcal{N}_e } e^{(\frac{  \varphi (h,r) \cdot \mathbf{t_j^-} }{\tau})}} \\
    
    = \frac{\varphi (h,r) }{\tau A} \sum\limits_{(h,r,t_j^-) \in \mathcal{N}_e } e^{(\frac{  \varphi (h,r) \cdot \mathbf{t_j^-} }{\tau})}

\end{array}
\label{appendix_gradienth+}
\end{equation}

\begin{equation}
\small
\begin{array}{l}
	-\frac{\partial S(h,r,t^{+})}{\partial t_j^-} 
	= \frac{\partial}{\partial t_j^-} (  \frac{\varphi (h,r) \cdot \mathbf{t^+}}{\tau}- 
	\log (e^{(\frac{\varphi (h,r) \cdot \mathbf{t_j^-}}{\tau})} \\ 
	 \ \ \ \ +\sum_{(h,r,t_j^-) \in \{p_e \mathcal{N}_e - (h,r,t_j^-)\} } e^{(\frac{  \varphi (h,r) \cdot \mathbf{t_j^-} }{\tau})} ) ) \\
	 
	 = \frac{- e^{(\frac{  \varphi (h,r) \cdot \mathbf{t_j^-} }{\tau})} \frac{\varphi (h,r)}{\tau} }{e^{(\frac{\varphi (h,r) \cdot \mathbf{t_j^-}}{\tau})} 
    +\sum\limits_{(h,r,t_j^-) \in \{ p_e \mathcal{N}_e - (h,r,t_j^-)\} } e^{(\frac{  \varphi (h,r) \cdot \mathbf{t_j^-} }{\tau})}} \\
    
    =-\frac{\varphi (h,r)}{\tau A} e^{(\frac{  \varphi (h,r) \cdot \mathbf{t_j^-} }{\tau})}
\end{array}
\label{appendix_gradienthj}
\end{equation} 
\paragraph{Gradients of Contrastive Relation}
Taking the \Cref{relation2} as an example, we give the detailed reasoning steps of gradient about $h$, $t$, $r^+$ and $r^{-}$.
\begin{equation}
\small
\begin{array}{l}
	-\frac{\partial S(h,r^{+},t)}{\partial h}
	
	= \frac{\partial}{\partial h} (  \frac{\varphi (h,r^+) \cdot \mathbf{t}}{\tau}- 
	\log (e^{(\frac{\varphi (h,r^+) \cdot \mathbf{t}}{\tau})} \\ 
	 \ \ \ \ +\sum_{(h,r_j^-,t) \in \mathcal{N}_r } e^{(\frac{  \varphi (h,r_j^-) \cdot \mathbf{t} }{\tau})} ) ) \\
	 
	 =  \frac{\varphi '(h,r^+) \mathbf{t} }{\tau}  - \\
	\ \ \ \ \frac{e^{(\frac{  \varphi (h,r^+) \cdot \mathbf{t} }{\tau})} \frac{\varphi '(h,r^+) \mathbf{t}}{\tau} + \sum\limits_{(h,r_j^-,t) \in \mathcal{N}_r } e^{(\frac{  \varphi (h,r_j^-) \cdot \mathbf{t} }{\tau})} \frac{\varphi '(h,r_j^-) \mathbf{t} }{\tau}  }{e^{(\frac{\varphi (h,r^+) \cdot \mathbf{t}}{\tau})} 
    +\sum\limits_{(h,r_j^-,t) \in \mathcal{N}_e } e^{(\frac{  \varphi (h,r_j^-) \cdot \mathbf{t} }{\tau})}} \\
    
    =  \frac{\varphi '(h,r^+) \mathbf{t} }{\tau B} \sum\limits_{(h,r_j^-,t) \in \mathcal{N}_r } e^{(\frac{  \varphi (h,r_j^-) \cdot \mathbf{t} }{\tau})}    - \\
	\ \ \ \ \frac{1}{B}\sum\limits_{(h,r_j^-,t) \in \mathcal{N}_r } e^{(\frac{  \varphi (h,r_j^-) \cdot \mathbf{t} }{\tau})} \frac{\varphi '(h,r_j^-) \mathbf{t} }{\tau}  \\
\end{array}
\label{appendix_gradientt}
\end{equation}

\begin{equation}
\small
\begin{array}{l}
	-\frac{\partial S(h,r^{+},t)}{\partial t}
	
	= \frac{\partial}{\partial t} (  \frac{\varphi (h,r^+) \cdot \mathbf{t}}{\tau}- 
	\log (e^{(\frac{\varphi (h,r^+) \cdot \mathbf{t}}{\tau})} \\ 
	 \ \ \ \ +\sum_{(h,r_j^-,t) \in \mathcal{N}_r } e^{(\frac{  \varphi (h,r_j^-) \cdot \mathbf{t} }{\tau})} ) ) \\
	 
	 =  \frac{\varphi (h,r^+) }{\tau}  - \\
	\ \ \ \ \frac{e^{(\frac{  \varphi (h,r^+) \cdot \mathbf{t} }{\tau})} \frac{\varphi (h,r^+) }{\tau} + \sum\limits_{(h,r_j^-,t) \in \mathcal{N}_r } e^{(\frac{  \varphi (h,r_j^-) \cdot \mathbf{t} }{\tau})} \frac{\varphi (h,r_j^-) }{\tau}  }{e^{(\frac{\varphi (h,r^+) \cdot \mathbf{t}}{\tau})} 
    +\sum\limits_{(h,r_j^-,t) \in \mathcal{N}_r } e^{(\frac{  \varphi (h,r_j^-) \cdot \mathbf{t} }{\tau})}} \\
    
    =  \frac{\varphi (h,r^+) }{\tau B} \sum\limits_{(h,r_j^-,t) \in \mathcal{N}_r } e^{(\frac{  \varphi (h,r_j^-) \cdot \mathbf{t} }{\tau})}  - \\
	\ \ \ \ \frac{1}{B}\sum\limits_{(h,r_j^-,t) \in \mathcal{N}_r } e^{(\frac{  \varphi (h,r_j^-) \cdot \mathbf{t} }{\tau})} \frac{\varphi (h,r_j^-) }{\tau}  \\
\end{array}
\end{equation}

\begin{equation}
\small
\begin{array}{l}
	-\frac{\partial S(h,r^{+},t)}{\partial r_j^{-}} \\
	
	= \frac{\partial}{\partial r_j^{-}} (  \frac{\varphi (h,r^+) \cdot \mathbf{t}}{\tau}- 
	\log ( e^{(\frac{\varphi (h,r_j^-) \cdot \mathbf{t}}{\tau})} \\ 
	 \ \ \ \ +\sum_{(h,r_j^{-},t) \in \{p_r \mathcal{N}_r - (h,r_j^{-},t)\} } e^{(\frac{  \varphi (h,r_j^{-}) \cdot \mathbf{t} }{\tau})} ) ) \\
	 
	= \frac{- e^{(\frac{  \varphi (h,r_j^-) \cdot \mathbf{t} }{\tau})} \frac{\varphi '(h,r_j^-) \mathbf{t}}{\tau} }
	{e^{(\frac{\varphi (h,r_j^-) \cdot \mathbf{t}}{\tau})}
    +\sum_{(h,r_j^{-},t) \in \{p_r \mathcal{N}_r - (h,r_j^{-},t) \} } e^{(\frac{  \varphi (h,r_j^{-}) \cdot \mathbf{t} }{\tau})}} \\
    
    = - \frac{\varphi '(h,r_j^-) \mathbf{t}}{\tau B}
	e^{(\frac{  \varphi (h,r_j^-) \cdot \mathbf{t} }{\tau})} \\
    
\end{array}
\label{appendix_gradienthrrr}
\end{equation}

\begin{equation}
\small
\begin{array}{l}
	-\frac{\partial S(h,r^{+},t)}{\partial r^{+}} 
	= \frac{\partial}{\partial r^{+}}( \log \frac{e^{(\beta(h,r^{+},t)/\tau)}}{\sum_{n \in \{p_r, \mathcal{N}_r\} } e^{(\beta(n)/\tau)}} ) \\
	= \frac{\partial}{\partial r^{+}}(  \frac{\beta(h,r^{+},t)}{\tau}- \log \sum_{n \in \{p_r, \mathcal{N}_r\} } e^{(\frac{\beta(n)}{\tau})} ) \\
	= \frac{\partial}{\partial r^{+}} (  \frac{\beta(h,r^{+},t)}{\tau}- 
	\log (e^{(\frac{\beta(h,r^{+},t)}{\tau})} \\ 
	 \ \ \ \ +\sum_{(h,r_j^{-},t) \in \mathcal{N}_r } e^{(\frac{\beta(h,r_j^{-},t)}{\tau})} ) ) \\
	
	= \frac{\partial}{\partial r^{+}} (  \frac{\varphi (h,r^+) \cdot \mathbf{t}}{\tau}- 
	\log (e^{(\frac{\varphi (h,r^+) \cdot \mathbf{t}}{\tau})} \\ 
	 \ \ \ \ +\sum_{(h,r_j^{-},t) \in \mathcal{N}_r } e^{(\frac{  \varphi (h,r_j^{-}) \cdot \mathbf{t} }{\tau})} ) ) \\
	 
	=  \frac{\varphi '(h,r^+) \mathbf{t}}{\tau}  - \\
	\ \ \ \ \frac{e^{(\frac{  \varphi (h,r^+) \cdot \mathbf{t} }{\tau})} \frac{\varphi '(h,r^+) \mathbf{t}}{\tau} }{e^{(\frac{\varphi (h,r^+) \cdot \mathbf{t}}{\tau})} 
    +\sum\limits_{(h,r_j^-,t) \in \mathcal{N}_r } e^{(\frac{  \varphi (h,r_j^-) \cdot \mathbf{t} }{\tau})}} \\
    
    =  \frac{\varphi '(h,r^+) \mathbf{t}}{\tau} ( \\
	\ \ \ \ \frac{\sum\limits_{(h,r_j^-,t) \in \mathcal{N}_r } e^{(\frac{  \varphi (h,r_j^-) \cdot \mathbf{t} }{\tau})}}{e^{(\frac{\varphi (h,r^+) \cdot \mathbf{t}}{\tau})} 
    +\sum\limits_{(h,r_j^-,t) \in \mathcal{N}_r } e^{(\frac{  \varphi (h,r_j^-) \cdot \mathbf{t} }{\tau})}} ) \\
    
    =  \frac{\varphi '(h,r^+) \mathbf{t}}{\tau B}
	\sum\limits_{(h,r_j^-,t) \in \mathcal{N}_r } e^{(\frac{  \varphi (h,r_j^-) \cdot \mathbf{t} }{\tau})}
\end{array}
\label{appendix_gradienth2}
\end{equation}

    

\begin{equation}
\small
\begin{array}{l}
B = e^{(\frac{\varphi (h,r^+) \cdot \mathbf{t}}{\tau})} 
    +\sum\limits_{(h,r_j^-,t) \in \mathcal{N}_r } e^{(\frac{  \varphi (h,r_j^-) \cdot \mathbf{t} }{\tau})}
\end{array}
\label{appendix_B}
\end{equation}

\begin{figure}
    \hspace{-3mm}
	\centering
	\includegraphics[width=1.05\columnwidth]{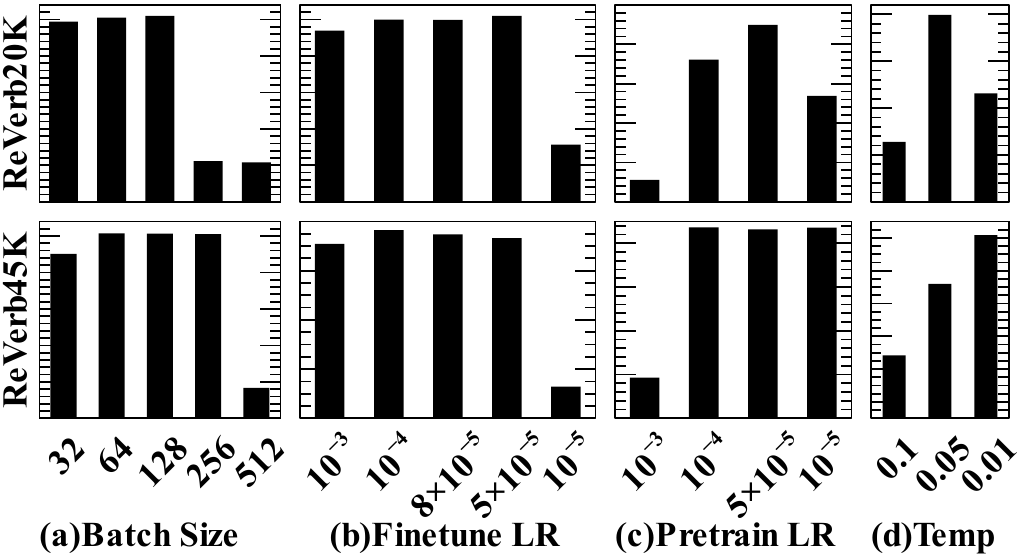}
	\caption{Results of hyperparameters on ReVerb20K (upper) and ReVerb45K (Below).}
	\label{all}
\end{figure}

\begin{table}[t]
\small
	\centering
	\resizebox{0.5\textwidth}{!}{
	\begin{tabular}{lcccccc}
		\toprule
		Model & $AR \downarrow$ & $ARR$ & $H@$1 & $H@$10 & $H@$50 & $H@$100 \\
		\midrule
        \emph{BERT-F} & 754 & 28.1 & 22.1 & 39.1 & 53.8 & 61.2 \\
        \emph{BERT-U} & 1315 & 23.7 & 18.7 & 33.2 & 45.9 & 51.8 \\
        \emph{BiGRU}  & \textbf{393} & \textbf{38.9} & \textbf{30.5} & \textbf{54.6} & \textbf{69.2} & \textbf{75.5} 
		\\
        \midrule
        \emph{BERT-F} & 1165 & 28.8 & 21.6 & 42.8 & 55.3 & 60.4 \\
        \emph{BERT-U} & 2131 & 18.0 & 12.5 & 28.9 & 41.3 & 46.6 \\
        \emph{BiGRU}  & \textbf{767} & \textbf{33.3} & \textbf{25.3} & \textbf{48.7} & \textbf{63.0} & \textbf{68.3} \\
		\bottomrule
	\end{tabular}
	}
	\caption{Results of Textual Technology on ReVerb20K (upper) and ReVerb45K (below).}
	\label{ab_b}
\end{table}

\subsection{Text Encoder}

Results of textual technologies to encoder sequences in \Cref{subsec:emb}: BiGRU and BERT, are shown in \Cref{ab}, where \emph{BERT-F} fixes the parameters of Bert, \emph{BERT-U} unfixes them.
By observing the results, performance of model with BERT (both fixed and unfixed parameters) as encoder is weaker than that with BiGRU. 
This shows that BiGRU is more capable of capturing sequential information in KGs, and pretrained language model can not give any help to TernaryCL.
State-of-the-art baseline OKGIT also points out that pretrained language models can not predict the correct entities on the top. It could be inadvisable to introduce the pretrained language model into the proposed model due to the difficulty of training and reproduction.
The intention of constructing a KG is to convert text into structures for easy calculation, that is, KG itself is rich in world knowledge.
So, our TernaryCL pays attention to mining own structure features in an effective way.

\end{document}